\documentclass{article}

\usepackage[preprint]{neurips_2023}

\usepackage[ruled,vlined]{algorithm2e}

\usepackage{enumerate}
\usepackage{amsmath,amsfonts}
\usepackage{graphicx}
\usepackage{url}

\begin{document}

\title{Using Linear Regression for Iteratively Training Neural Networks}

\author{%
  Harshad~Khadilkar\thanks{Also visiting associate professor at IIT Bombay, harshadk@cse.iitb.ac.in} \\
  TCS Research\\
  Mumbai, India \\
  \texttt{harshad.khadilkar@tcs.com} \\
}

\maketitle

\begin{abstract}
  We present a simple linear regression based approach for learning the weights and biases of a neural network, as an alternative to standard gradient based backpropagation. The present work is exploratory in nature, and we restrict the description and experiments to (i) simple feedforward neural networks, (ii) scalar (single output) regression problems, and (iii) invertible activation functions. However, the approach is intended to be extensible to larger, more complex architectures. The key idea is the observation that the input to every neuron in a neural network is a linear combination of the activations of neurons in the previous layer, as well as the parameters (weights and biases) of the layer. If we are able to compute the ideal total input values to every neuron by working backwards from the output, we can formulate the learning problem as a linear least squares problem which iterates between updating the parameters and the activation values. We present an explicit algorithm that implements this idea, and we show that (at least for small problems) the approach is more stable and faster than gradient-based methods.
\end{abstract}



\section{Introduction}

The training of neural networks is known to be a hard problem \citep{glorot2010understanding}, and several strategies have been proposed to tackle it. The standard backpropagation strategy is known to have empirically poor results \citep{larochelle2009exploring}, leading to the following variations. 

\subsection{Related work}

Modifications for stabilising backpropagation include (i) initialisation strategies such as the use of complementary priors \citep{hinton2006fast} or the stacking of Restricted Boltzmann Machines for weight initialisation \citep{hinton2007recognize}, (ii) unsupervised approaches that aim to preserve information transfer through successive layers \citep{larochelle2009exploring,lowe2019putting}, and (iii) regularisation and dropout \citep{wan2013regularization,zaremba2014recurrent,srivastava2014dropout}. Nevertheless, training using gradient descent from random initialisation is known to be brittle \citep{glorot2010understanding} and possibly not a good model of biological learning \citep{lillicrap2020backpropagation,hinton2022forward}.

Recently, the forward-forward algorithm \citep{hinton2022forward} was proposed for training neural networks in a new way that increases the weights for positive samples and decreases them for negative samples. However, this algorithm is expected to outperform backpropagation mainly in power-constrained or uncertain computational settings. An even more recent approach called Cascaded Forward \citep{zhao2023cascaded} aims to improve on this by forcing each layer to successfully represent the output probability distribution. The disadvantage of such approaches is that they move quite far from the basic optimisation objective (output error minimisation) by imposing theoretical interpretations of weights, biases, and biological processes.

An alternative to theoretically complex approaches is the intuitively simple concept of neuroevolution \citep{stanley2019designing,mirjalili2019evolutionary}. This idea proposes to do away with gradient based training, and instead implements randomised search on a large population of parameter vectors to find the strongest (most optimal) ones. While easy to understand and implement, neuroevolution is computationally prohibitive for training large neural networks.

\subsection{Contributions}

Instead, we propose to use the linearisation of training in neural networks directly for parameter optimisation, as described in Sec. \ref{sec:method}. The method is inspired by two ideas from prior literature. The first idea is the use of orthogonal least squares for the training of radial basis functions (RBF) \citep{chen1991orthogonal,huang2005determining}. The output of an RBF is observed to be composed of a linear combination of basis vectors, followed by a non-linearity. This observation is then used to compute the centers of new hidden units. The second idea is the point-wise linearity of training dynamics in wide neural networks \citep{lee2019wide}. This study notes that in the infinite-width limit, the training dynamics of a neural network can be described using Taylor series expansions of the initial parameter values. Based on this inspiration, we claim the following contributions for this paper.
\begin{enumerate}[a.]
\item An algorithm that uses iterative linear least squares (ILLS) that uses the inherent linear relationship between the output of one layer and the input of the next layer, to provide an alternative training method for neural networks. In this early work, we restrict the description to feedforward networks with invertible activation functions. 
\item A set of experiments on 1-layer and 2-layer multi-input neural networks, that demonstrate a significant advantage over standard backpropagation using adaptive moment estimation. The experiments include 3 synthetic data sets and one publicly available data set.
\end{enumerate}

\section{Solution Methodology} \label{sec:method}

\subsection{Illustrative network}

\begin{figure}
\centering
\includegraphics[width=0.5\textwidth]{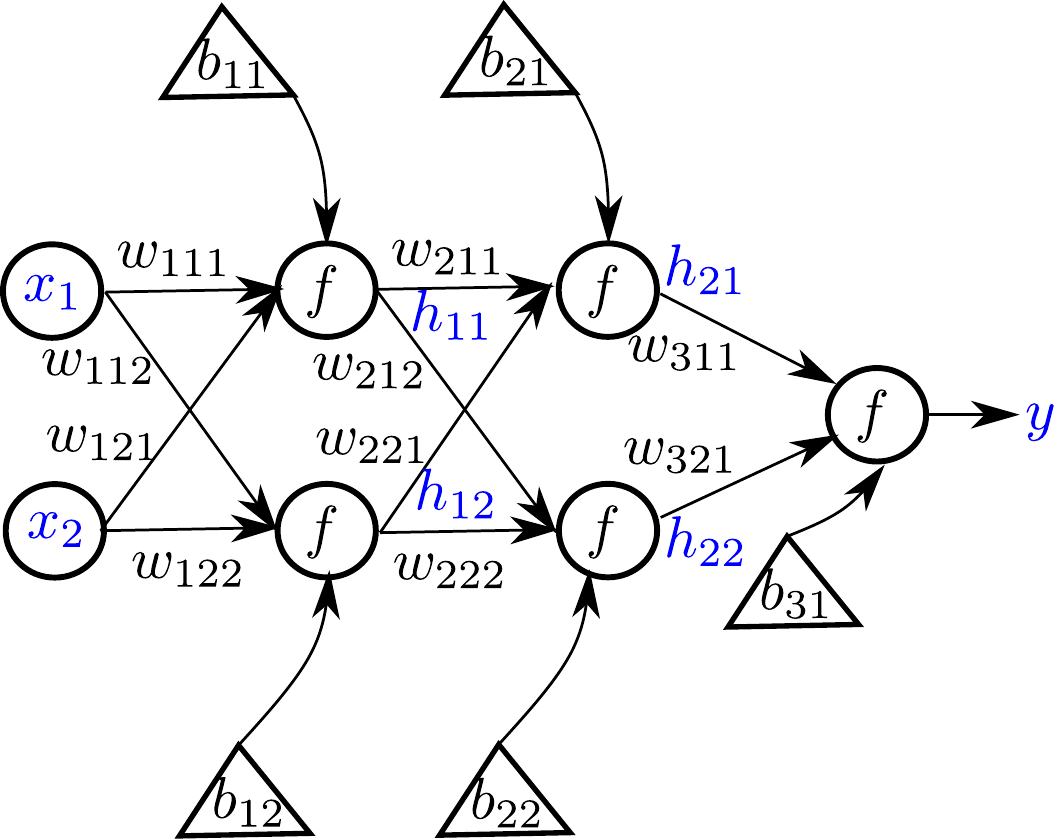}
\caption{Sample network with two hidden layers.}
\label{fig:network2}
\end{figure}

Consider the neural network shown in Figure \ref{fig:network2}. The two inputs $x_1$ and $x_2$ pass through two hidden layers with two neurons each, and produce a single output ${y}$. The activation of each neuron is $f$, and is assumed to be invertible (one-to-one mapping from input to output). The goal of training is to compute the values of weights $w_{ijk}$ and biases $b_{ij}$ that minimise the mean-squared error between the predicted and the ideal output for each training data sample. In this notation, $i$ is the index of the layer, $j$ is the index of neuron within the layer, and $k$ is the index of the neuron in the next layer. The input-output relationships are given by,
\begin{align}
h_{11} & = f(w_{111}\,x_1 + w_{121}\,x_2 - b_{11}) \label{eq:h11} \\
h_{12} & = f(w_{112}\,x_1 + w_{122}\,x_2 - b_{12}) \label{eq:h12} \\
h_{21} & = f(w_{211}\,h_{11} + w_{221}\,h_{12} - b_{21}) \label{eq:h21} \\
h_{22} & = f(w_{212}\,h_{11} + w_{222}\,h_{12} - b_{22}) \label{eq:h22} \\
{y} & = f(w_{311}\,h_{21} + w_{321}\,h_{22} - b_{31}) \label{eq:y}
\end{align}

\subsection{Preliminaries}

Consider the network architecture shown in Figure \ref{fig:network2}. In the following description, we denote the estimated values of parameters by hat superscripts. For example, the estimated value of $w_{111}$ is written as $\hat{w}_{111}$. We also denote the training data set of $N$ samples by tuples $(x_1,x_2,y)_n$, where $n\in\mathbb{Z}^+$ and $1\leq n \leq N$. Note that we have $N$ simultaneous equations to be satisfied. At the same time, we have $(4N+15)$ unknowns consisting of $15$ network parameters and $4N$ hidden variables. The system is thus underdetermined. We now consider the output equation \eqref{eq:y}. Since $f$ is assumed to be invertible, we can directly compute the vector $H=f^{-1}(y)$ for the data set, which is the total input required for the output neuron. The following relationship must hold if the output error is $0$,
\begin{equation}
\hat{w}_{311}\,\hat{h}_{21} + \hat{w}_{321}\,\hat{h}_{22} - \hat{b}_{31} = H.
\label{eq:H}
\end{equation}
We note that \eqref{eq:H} is linear in $(\hat{h}_{21},\hat{h}_{22})$ when $(\hat{w}_{311},\hat{w}_{321},\hat{b}_{31})$ are held constant, and vice versa. We may therefore consider solving this equation using iterative linear squares. The first step with constant $(\hat{h}_{21},\hat{h}_{22})$ is straightforward. Given current values of the hidden activations, \eqref{eq:H} is a system of $N$ linear equations which can be solved to find the optimal values of $(\hat{w}_{311},\hat{w}_{321},\hat{b}_{31})$. The next step is trickier. For the updated values of $(\hat{w}_{311},\hat{w}_{321},\hat{b}_{31})$, we may consider solving for the new values of $(\hat{h}_{21},\hat{h}_{22})$. However, there are $2N$ of these variables, and we cannot arbitrarily vary the hidden activation values since these must be the outputs of the previous set of neurons. We therefore propose to take a small step along the linearised gradient of the activation function $f$. Consider the hidden activation $\hat{h}_{21}$. The input to the corresponding neuron is given using \eqref{eq:h21} as,
\begin{equation}
I_{21} = \hat{w}_{211}\,\hat{h}_{11} + \hat{w}_{221}\,\hat{h}_{12} - \hat{b}_{21} = f^{-1}(\hat{h}_{21})
\label{eq:I21}
\end{equation}
Since we propose to update the weight and bias parameters of the previous layer in order to modify $\hat{h}_{21}$, we need the partial derivatives of $I_{21}$ with respect to the parameters,
\begin{equation}
\frac{\partial I_{21}}{\partial \hat{w}_{211}} = \hat{h}_{11},\; \frac{\partial I_{21}}{\partial \hat{w}_{221}} = \hat{h}_{12},\; \frac{\partial I_{21}}{\partial \hat{b}_{21}} = -1
\label{eq:partials}
\end{equation}
Using \eqref{eq:h21}, the linearised change in activation $\hat{h}_{21}$ given small deviations to the previous parameters is,
\begin{equation}
\Delta \hat{h}_{21} = \frac{\partial f}{\partial I_{21}} \Bigr|_{{\hat{h}_{11},\hat{h}_{12}}} \cdot \left( \alpha_{211}\,\hat{h}_{11} + \alpha_{221}\,\hat{h}_{12} - \beta_{21} \right),
\label{eq:deltah21}
\end{equation}
where $\alpha_{211},\alpha_{221},\beta_{21}$ are the proposed deviations in the values of $\hat{w}_{211},\hat{w}_{221},\hat{b}_{21}$ respectively. Along similar lines, we can also write
\begin{equation}
\Delta \hat{h}_{22} = \frac{\partial f}{\partial I_{22}} \Bigr|_{{\hat{h}_{11},\hat{h}_{12}}} \cdot \left( \alpha_{212}\,\hat{h}_{11} + \alpha_{222}\,\hat{h}_{12} - \beta_{21} \right).
\label{eq:deltah22}
\end{equation}
Instead of solving \eqref{eq:H} directly to get $\hat{h}_{21},\hat{h}_{22}$, we solve for the deviation in the previous layer parameters, which lets us (approximately) satisfy the input-output relationship of the neuron. We thus solve,
\begin{equation*}
\hat{w}_{311}\,(\hat{h}_{21}+\Delta \hat{h}_{21}) + \hat{w}_{321}\,(\hat{h}_{22}+\Delta \hat{h}_{22}) - \hat{b}_{31} = H
\end{equation*}
\begin{equation}
\Rightarrow \hat{w}_{311}\,\Delta \hat{h}_{21} + \hat{w}_{321}\,\Delta \hat{h}_{22} = H + \hat{b}_{31} - \hat{w}_{311}\,\hat{h}_{21} - \hat{w}_{321}\,\hat{h}_{22},
\label{eq:linearised}
\end{equation}
where $\Delta \hat{h}_{21},\,\Delta \hat{h}_{22}$ are given by \eqref{eq:deltah21} and \eqref{eq:deltah22}. Note that \eqref{eq:linearised} is a linear equation in 6 variables $\alpha_{211},\alpha_{221},\beta_{21},\alpha_{212},\alpha_{222},\beta_{22}$. This can be solved to compute proposed deviations in the weights and biases of the previous layer (which are unconstrained and small in number), rather than \eqref{eq:H}, which would compute deviations in the constrained hidden activations which are $2N$ in number. Since this is a linear approximation, it is only valid in a small region around $\hat{h}_{21},\,\hat{h}_{22}$. Therefore we define a learning rate $\rho$ for updating the values,
\begin{equation}
\hat{h}_{21}\leftarrow \hat{h}_{21}+\rho\,\frac{\Delta \hat{h}_{21}}{N(\Delta \hat{h}_{21},\Delta \hat{h}_{22})},\;\hat{h}_{22}\leftarrow \hat{h}_{22}+\rho\,\frac{\Delta \hat{h}_{22}}{N(\Delta \hat{h}_{21},\Delta \hat{h}_{22})},
\label{eq:update}
\end{equation}
where $N(\Delta \hat{h}_{21},\Delta \hat{h}_{22})$ is a common normalisation constant that maps the un-normalised values of $\Delta \hat{h}_{21},\Delta \hat{h}_{22}$ to the range $[-1,1]^N$. We emphasise that the deviations in the hidden activations are computed based on the regressed values of $\alpha$'s and $\beta$'s, and thus correspond to approximately feasible updates for the parameters.

\begin{algorithm}[b]
	\SetAlgoLined
	\KwIn{Initial guesses for weights $\hat{w}_{ijk}$ and biases $\hat{b}_{ij}$, training data set of $N$ tuples $(x_1,x_2,y)_n$, where $n\in\mathbb{Z}^+,\,1\leq n\leq N$, learning rate $\rho$.}
	\For{$e \gets 1$ \textbf{to} $\max\_\mathrm{epochs}$} 
	{
		i. Compute forward pass using \eqref{eq:h11}-\eqref{eq:y}, and estimates of $\hat{w}_{ijk}$ and $\hat{b}_{ij}$ instead of true (unknown) values \\
		ii. Update $(\hat{w}_{311},\hat{w}_{321},\hat{b}_{31})$ using \eqref{eq:H}, constant $\hat{h}_{21},\hat{h}_{22}$ \\
		iii. Compute $\alpha_{211},\alpha_{221},\beta_{21},\alpha_{212},\alpha_{222},\beta_{22}$ using \eqref{eq:linearised} \\
		iv. Update $\hat{h}_{21}$ and $\hat{h}_{22}$ using \eqref{eq:update} \\
		v. Compute $(\hat{w}_{211},\hat{w}_{221}, \hat{b}_{21})$ by fitting \eqref{eq:I21} to $f^{-1}(\hat{h}_{21})$ \\
		vi. Compute $(\hat{w}_{212},\hat{w}_{222}, \hat{b}_{22})$ by fitting $I_{22}$ to $f^{-1}(\hat{h}_{22})$ \\
		vii. Update $\hat{h}_{11}$ and $\hat{h}_{12}$ by computing $\alpha_{111},\alpha_{121},\beta_{11},$ $\alpha_{112},\alpha_{122},\beta_{12}$ that minimise error with respect to $I_{21}$ \\
		viii. Update $\hat{h}_{11}$ and $\hat{h}_{12}$ by computing $\alpha_{111},\alpha_{121},\beta_{11},$ $\alpha_{112},\alpha_{122},\beta_{12}$ that minimise error with respect to $I_{22}$ \\
		ix. Compute $\hat{w}_{111},\hat{w}_{121},\hat{b}_{11}$, minimise error to $f^{-1}(\hat{h}_{11})$ \\
		x. Compute $\hat{w}_{112},\hat{w}_{122},\hat{b}_{12}$, minimise error to $f^{-1}(\hat{h}_{12})$ \\
	}
	\Return{final values of $\hat{w}_{ijk}$ and $\hat{b}_{ij}$}
	\caption{Training of the 2-layer network from Fig. \ref{fig:network2}.}
	\label{algo1}
\end{algorithm}

Taken together, the computation of $(\hat{w}_{311},\hat{w}_{321},\hat{b}_{31})$ using \eqref{eq:H} and then the updation of $\hat{h}_{21},\hat{h}_{22}$ using \eqref{eq:update} constitutes one batch update of the last layer of the network from Figure \ref{fig:network2}. We utilise the same logic to proceed backwards up to the input layer, with two small modifications in the following cases (details in Algorithm \ref{algo1}).
\begin{enumerate}[a.]
\item When a hidden activation is sent to multiple neurons in the next layer, we make one update for every destination neuron. For example, $\hat{h}_{11}$ and $\hat{h}_{12}$ in Figure \ref{fig:network2} drive both $\hat{h}_{21}$ and $\hat{h}_{22}$. We can thus compute one update analogous to \eqref{eq:update} via $\hat{h}_{21}$, and another via $\hat{h}_{22}$. In Algorithm \ref{algo1}, we make one update to $\hat{h}_{11}$ and $\hat{h}_{12}$ (and hence to the four weights and two biases in the input layer) based on $\hat{h}_{21}$, and another based on $\hat{h}_{22}$ in every backward pass. 
\item In the final step of the backward pass (at the input layer), we only update the weights and biases, since the inputs $x_1$ and $x_2$ are known and fixed.
\end{enumerate}

\section{Results} \label{sec:results}

\subsection{A note on initialisation}

Algorithm \ref{algo1} shows that the initial values of $\hat{w}_{ijk}$ and $\hat{b}_{ij}$ have an important effect on training, since they affect the initial values of hidden activations which result in the first update in step (ii). As a result, the experiments reported below (all of which use \texttt{tanh} activation) use two types of initialisation. The first `default' option is using the standard initialiser in \texttt{PyTorch}. The second `custom' initialisation uses the following distributions for weights and biases,
\begin{align*}
\hat{w}_{ijk} & \sim \mathcal{U} \left( [-0.75,-0.25] \cup [0.25,0.75] \right) \\
\hat{b}_{ij} & \sim \mathcal{U} [-0.1,0.1],
\end{align*}
where $\mathcal{U}$ is the uniform distribution. The ranges defined above ensure that the \texttt{tanh} activation in every layer covers a substantial portion of the $[-1,1]$ output range, and also has sufficiently large magnitudes for the partial derivatives in \eqref{eq:partials}.

\subsection{Experiments}

We run four experiments to demonstrate the effectiveness of the ILLS algorithm, as described below. All experiments use \texttt{tanh} activation for all neurons, since it satisfies the invertibility requirement on $f$. We use the \texttt{Adam} optimiser in \texttt{PyTorch} as the baseline. Both algorithms (ILLS and gradient descent) are run with both the default and the custom initialisations described above. For any given random seed, both algorithms start from identical initial values of weights and biases for the same type of initialisation. All results are reported using mean and standard deviation on 10 random seeds.

\begin{figure}[!h]
\centering
\includegraphics[width=0.78\textwidth]{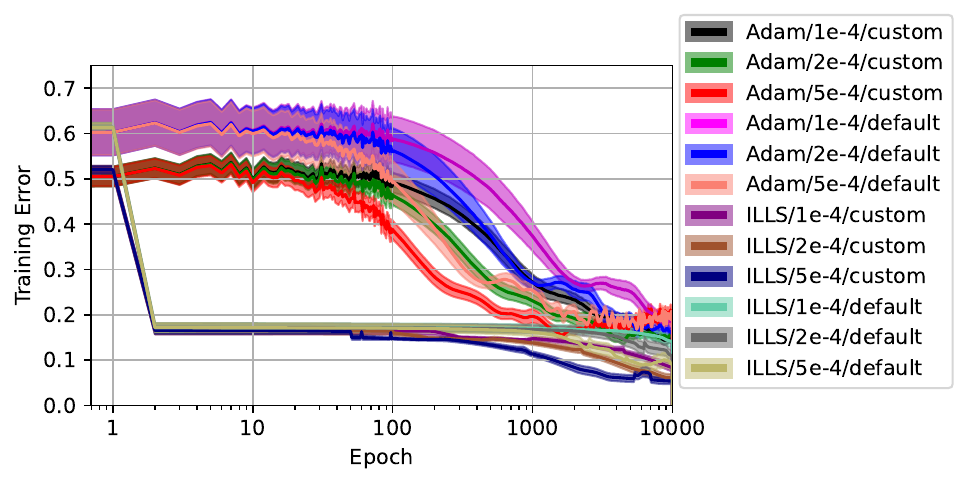}
\caption{Comparison of training loss for Adam and ILLS, for the two-layer network with assumed parameters $w_{111}=-0.5, w_{121}=0, w_{112}=-1, w_{122}=1, b_{11}=1, b_{12}=-1, w_{211}=2, w_{212}=-0.5, w_{221}=0.5, w_{222}=2, b_{21}=0, b_{22}=2, w_{311}=1, w_{321}=-1, b_{31}=-1$. Curves correspond to different learning rates and initialisation schemes. Shaded regions show standard deviation over 10 random seeds. Note that Adam starts to diverge at the highest learning rates shown, and thus cannot be further increased.}
\label{fig:set1_two_layer}
\end{figure}

\textbf{Experiment 1 (synthetic, 2 layer):} Considering the network shown in Figure \ref{fig:network2}, we generate a data set of $N=100$ pairs $(x_1,x_2)$ from a uniform distribution on $[-1,1]$. Using a synthetically generated set of weight and bias values, we compute the ground truth output $y_n$ for each sample. This set is fed to Algorithm \ref{algo1} with different values of learning rate $\rho$. The same network is also implemented using \texttt{PyTorch} and trained using the \texttt{Adam} optimiser on the same data set, with different learning rates and default momentum value. A randomly shuffled version of the entire data set is used as the batch for every epoch. Figure \ref{fig:set1_two_layer} shows the comparison results averaged over 10 random seeds for each algorithm, with both default and custom initialisation. The x-axis is logarithmic. Starting from the same initial training error, ILLS takes a large downward step in the first iteration, and ends with a lower final error than \texttt{Adam}. The custom initialisation starts with a lower error for both algorithms, though this is probably incidental. Finally, we note that \texttt{Adam} begins to diverge for the highest learning rates in the figure ($5\times 10^{-4}$), possibly due to overfitting on the small sized data set. This behaviour is not observed for ILLS.

\begin{figure}[!h]
\centering
\includegraphics[width=0.28\textwidth]{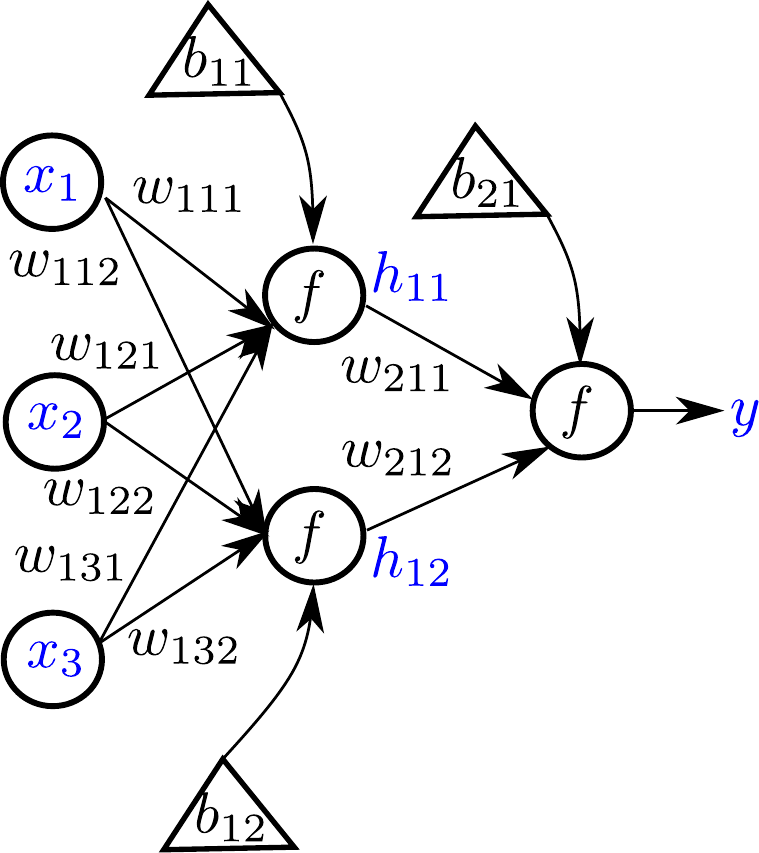}
\caption{Network with one hidden layer, used experiments 2, 3, and 4.}
\label{fig:network1}
\end{figure}

\begin{figure}[!h]
\centering
\includegraphics[width=0.78\textwidth]{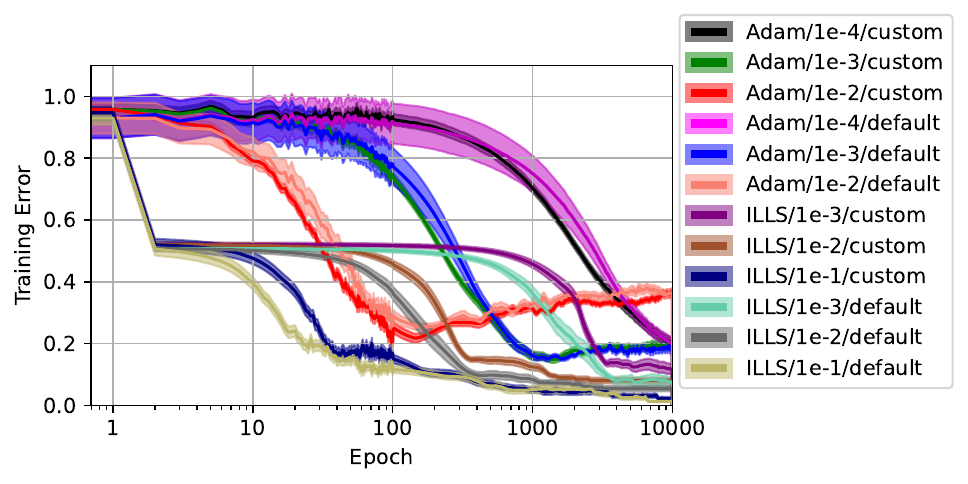}
\caption{Comparison of training loss for Adam and ILLS, for the one-layer network with assumed parameters $w_{111}=1, w_{121}=2, w_{131}=-1, w_{112}=0, w_{122}=3, w_{132}=-2, b_{11}=1, b_{12}=-2, w_{211}=3, w_{221}=-2, b_{21}=0$. Curves correspond to different learning rates and initialisation schemes. Shaded regions show standard deviation over 10 random seeds. Note that Adam starts to diverge at the highest learning rates shown, and thus cannot be further increased.}
\label{fig:set2_one_layer}
\end{figure}

\begin{figure}[!h]
\centering
\includegraphics[width=0.78\textwidth]{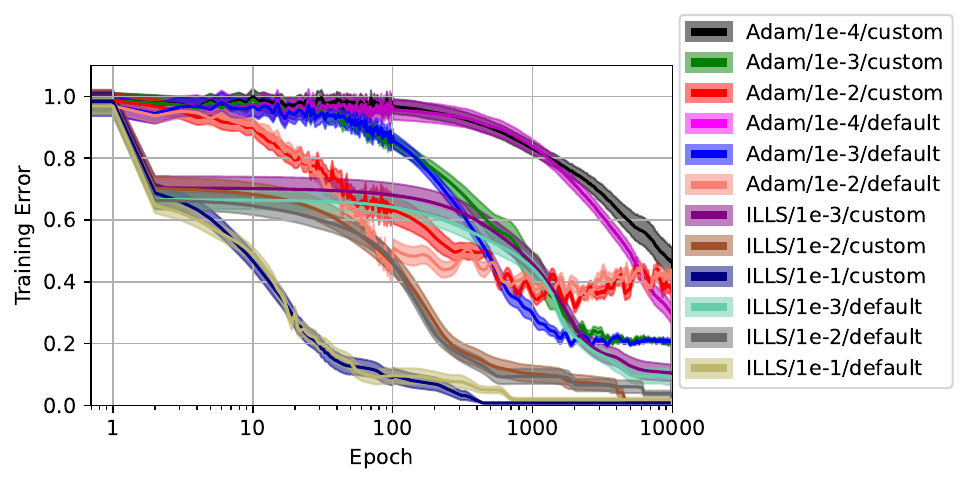}
\caption{Comparison of training loss for Adam and ILLS, for the one-layer network with assumed parameters $w_{111}=-2, w_{121}=0, w_{131}=3, w_{112}=4, w_{122}=-1, w_{132}=2, b_{11}=0, b_{12}=2, w_{211}=-3, w_{221}=-1, b_{21}=-1$. Curves correspond to different learning rates and initialisation schemes. Shaded regions show standard deviation over 10 random seeds. Note that Adam starts to diverge at the highest learning rates shown, and thus cannot be further increased.}
\label{fig:set3_one_layer}
\end{figure}

\textbf{Experiments 2 and 3 (synthetic, 1 layer):} In this pair of experiments, we use the network with one hidden layer and 3 inputs as shown in Figure \ref{fig:network1}. The update relationships can be written down by following the procedure outlined in Sec. \ref{sec:method}, but adapted to three inputs instead of two. Specifically, the linearised deviation to $\hat{h}_{11},\hat{h}_{12}$ in Figure \ref{fig:network1} contains six $\alpha$'s (corresponding to the six weights in the input layer) and two $\beta$'s, instead of four $\alpha$'s and two $\beta$'s. Figures \ref{fig:set2_one_layer} and \ref{fig:set3_one_layer} show a comparison of the training performance for two different sets of ideal weights and biases (specified in the respective captions). In both cases, we see a similar behaviour to that observed in Experiment 1. ILLS takes a large initial step followed by continuous improvement, with higher learning rates converging earlier. This time, we are not able to run \texttt{Adam} on the fastest rate of $0.1$ because it begins to diverge at a rate of $0.01$ itself. 

\begin{figure}[!h]
\centering
\includegraphics[width=0.7\textwidth]{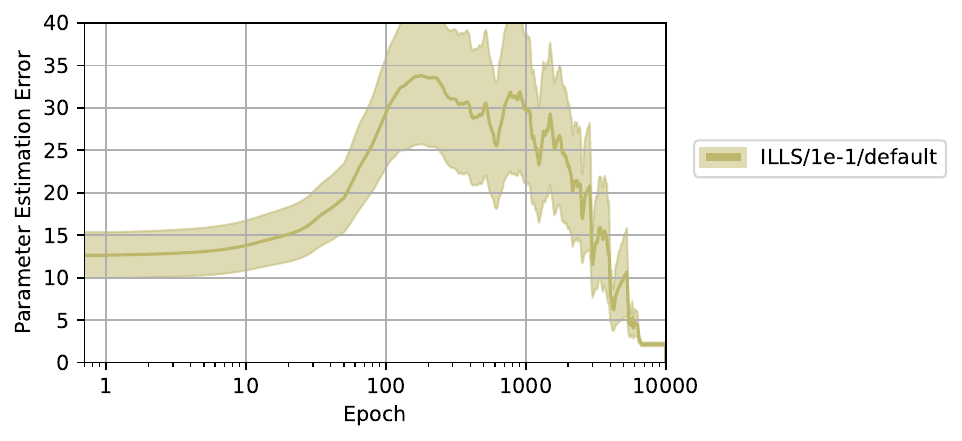}
\caption{Plot of $L2$ norm error between true and estimated parameters, for experiment 2 averaged over 10 random seeds, with the ILLS algorithm and a learning rate of $\rho=0.1$. We plot the error between absolute values of true and estimated parameters, for reasons explained in the text.}
\label{fig:parameters}
\end{figure}

From Figure \ref{fig:parameters}, we can also confirm that ILLS not only minimises the output error, but is able to compute a good approximation of the true parameters.  Because of the symmetry of \texttt{tanh} along both input and output axes, the optimal set of weights and biases come with an equally optimal solution with opposite sign. Specifically, we observe that,
\begin{equation*}
\mathrm{tanh}\left(\sum w_{ijk} \,h_{ijk} - b_{ij}\right) = -\mathrm{tanh} \left(\sum (-w_{ijk})\,h_{ijk} - (-b_{ij})\right),
\end{equation*}
which can be mapped to the required value by also flipping the sign of the weight of the next layer. As a result, Figure \ref{fig:parameters} plots the error in element-wise absolute values of the parameter estimates.

\begin{figure}[!h]
\centering
\includegraphics[width=0.78\textwidth]{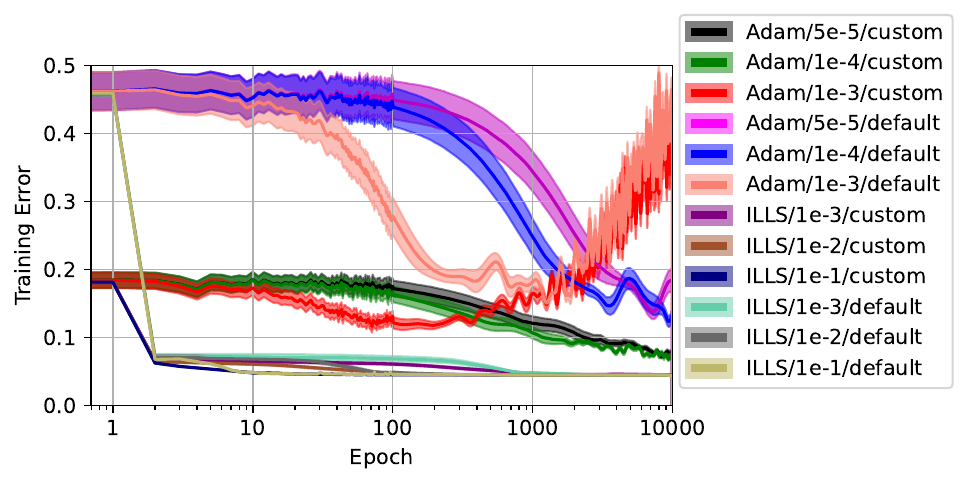}
\caption{Comparison of training loss for Adam and ILLS, for the one-layer network on the airline passenger data set. Curves correspond to different learning rates and initialisation schemes. Shaded regions show standard deviation over 10 random seeds. Note that Adam diverges at the highest learning rates shown, and thus cannot be further increased.}
\label{fig:airline_training}
\end{figure}

\textbf{Experiment 4 (public data set, 1 layer network): } For our final experiment, we use the US airline passengers data set \citep{airpassengers} available by default in \texttt{R-v3.6}. This data contains a time series of monthly passenger counts flown by airlines in the US between 1949 and 1960. For the experiment, we normalise the time series and create a training data set with 3 successive time steps as the input and the fourth step as the expected output, resulting in 142 data samples. The network shown in Figure \ref{fig:network1} is used for modelling the problem. Figure \ref{fig:airline_training} shows the comparison as before, with different learning rates. Note that in this case Adam begins to diverge at a rate of $0.001$ itself for both types of initialisation. ILLS is able to converge for rates at least as high as $0.1$, even though the ideal parameters in this case are unknown.

\newpage

\section{Discussion}

The description of ILLS in Sec. \ref{sec:method} and the experiments in Sec. \ref{sec:results} are exploratory in nature, and are not meant to be conclusive evidence for ILLS being better than backpropagation. Let us consider a few important open questions based on this thought.
\begin{enumerate}[a.]
\item \textbf{Is ILLS just gradient descent in another form?} A quick glance at the partial derivatives in \eqref{eq:partials} and the step update in \eqref{eq:update} raise the possibility of ILLS being equivalent to gradient descent. However, this is not the case because of the following reasons. 

	\begin{enumerate}[i.]
	\item Backpropagation with gradient descent effectively involves computing the sensitivity of the output error with respect to every parameter, and then taking a step opposite to the direction of the gradient. This is an `open loop' update; once the gradients are computed, every parameter is updated independently. By contrast, ILLS computes the linearised optimal value of the parameters and the hidden activations in an alternating fashion. The parameter values are directly updated without any need for a learning rate. Only the hidden activations are updated in small steps, in recognition of their relationship to the weights and biases in the previous layer. Furthermore, this is a `closed loop' update; the updates for every layer are designed to minimise the error to the \textit{updated} activations and parameters of the next layer.
	\item We do not need to handle any chain of non-linearities (over multiple layers) when using ILLS. The only derivatives appear in constant multipliers in relations such as \eqref{eq:deltah21} and \eqref{eq:deltah22}, and are limited to a single non-linearity. In essence, we solve a hierarchy of compact optimisation problems, starting from the final layer and proceeding backwards. The parameters in some arbitrary layer $i$ are not concerned with the error gradient with respect to the output; they are only focused on minimising the error to the values in layer ($i+1$).
	\item From the previous two points, it is clear that ILLS cannot be easily parallelised, since the updates have to be cascaded backwards. It may be possible to do a staggered parallelised update (with one iteration delay in successive layers), but this is yet to be explored.
	\end{enumerate}
\item \textbf{How will ILLS extend to other activation functions?} The description provided in this paper is valid for all invertible activation functions such as \texttt{sigmoid}, \texttt{tanh}, leaky ReLU, etc. We have not discussed the applicability of ILLS to non-invertible functions. In this case, we note that while there are common activation functions with a many-to-one relationship from input to output (for example, the negative portion of ReLU), almost all activations have a substantial strictly monotonic region (such as the positive part of a ReLU). We may be able to apply ILLS to subsets of the training data which correspond to only this portion of the activation. Recall that given the current parameter estimates, the forward pass of the network can be carried out for every data point. In the update step, and for every individual layer, we can identify the subset of data that correspond to monotonic portions of the activation for all neurons in that layer. Once the regression is solved for this subset of data, the updated values of weights and biases are valid for the entire data set. However, this proposal is untested at the moment.
\item \textbf{How will ILLS extend to other loss functions?} There are two types to be considered.
	\begin{enumerate}[i.]
	\item General $L^p$ losses: The description in this paper covers the standard $L^2$ loss which is used in linear least squares regression. In order to minimise generic $L^p$ loss, there is no structural change required to the algorithm; one can simply use the relevant linear best-fit optimiser to solve for the parameters in the final layer. In all previous layers, we can continue to use the least-squares formulation to match the required hidden activation values.
	\item Cross-entropy loss: In this case, the `ideal' outputs $y$ are one-hot vectors, which are not useful for ILLS because the ideal input values correspond to $\pm \infty$. It may be possible to approximate the one-hot vector by a softer version (for example, by replacing output value of $1$ by $0.99$) and then using ILLS. Alternatively, one could translate outputs $0$ to $-0.5$ and $1$ to $+0.5$ respectively. However, these are speculative ideas. The most promising idea\footnote{Suggested by Indrajit Bhattacharya, TCS Research} is to solve the final layer parameter estimation problem as a logistic regression problem (assuming $f$ is the sigmoid function for classification tasks). All previous layers can continue to operate as $L^2$ regression, since they only need to match the subsequent hidden activations.
	\end{enumerate}
\item \textbf{Will ILLS scale to large networks?} Computationally, ILLS does not appear to have any higher overheads than standard gradient based backpropagation. The mathematical operations required for gradient computation are replaced by the ones for computing the coefficients of linear regression. However, the numerical stability of the procedure is untested at this time. We also reiterate from remark (a.iii.) above, that parallelising the updates for ILLS might be a challenge.
\item \textbf{What about convergence of ILLS?} Step (ii) in Algorithm \ref{algo1} is a contraction mapping, since it directly solves a linear regression problem. So long as the linear approximation in step (iv) is valid, the update \eqref{eq:update} should also lead to a reduction in the residual. Therefore we believe ILLS to have stable convergence, although we do not have a rigorous proof at this time.
\end{enumerate}
In conclusion, ILLS is a mathematical curiosity at the present time. However, it appears to have potential to make an impact in the training tools available for deep learning. There is plenty of work to be done, in terms of analysis as a hierarchical optimisation problem, engineering to make it stable across a range of scales and problems, extensions to different activations and loss functions, and many other directions.


\bibliographystyle{ACM-Reference-Format}
\bibliography{refs}

\end{document}